\documentclass{llncs}
\usepackage{amssymb}
\usepackage{graphicx}
\usepackage{multirow}
\usepackage{amsmath}
\usepackage{pgfplots}
\usepackage{tikz}
\usepackage{url}
\usepackage{hyperref}
\usepackage[T2A]{fontenc}
\usepackage[utf8]{inputenc}
\usepackage{graphicx}
\usepackage{amsfonts}
\usepackage{multicol}
\usepackage{subfigure}
\usepackage{mathrsfs}
\usepackage{setspace}
\usepackage{lineno}
\usepackage{lscape}

\newcommand*{\Prob}{\mathsf{P}}
\newcommand{\softmax}{\mathop{\rm softmax}}
\newcommand{\TT}{\mathop{\rm TT}}

\begin{document}

\mainmatter  

\title{Neural Networks Compression for Language Modeling}

\titlerunning{Neural Networks Compression for Language Modeling}

%
%
\author{Artem~M.~Grachev\inst{1, 2} %
\and  Dmitry~I. Ignatov \inst{2} 
\and Andrey~V. Savchenko \inst{3} }  %

\institute{
Samsung R\&D Institute Rus, Moscow, Russia \\
\and
National Research University Higher School of Economics, Moscow, Russia \\
\and
National Research University Higher School of Economics, Laboratory of Algorithms and Technologies for Network Analysis, Nizhny Novgorod, Russia \\ 
\email{grachev.art@gmail.com}
}

\tocauthor{Authors' Instructions}
\maketitle

\begin{abstract}

In this paper, we consider several compression techniques for the language modeling problem based on recurrent neural networks (RNNs). It is known that conventional RNNs, e.g, LSTM-based networks in language modeling, are characterized with either high space complexity or substantial inference time. This problem is especially crucial for mobile applications, in which the constant interaction with the remote server is inappropriate. By using the Penn Treebank (PTB) dataset we compare pruning, quantization, 
low-rank factorization, tensor train decomposition for LSTM networks in terms of model size and suitability for fast inference.

\keywords{LSTM, RNN, language modeling, low-rank factorization, pruning, quantization}

\end{abstract}

\section{Introduction}

Neural network models can require a lot of space on disk and in memory. They can also need a substantial amount of time for inference.  
This is especially important for models that we put on devices like mobile phones. 
There are several approaches to solve these problems.
Some of them are based on sparse computations. They also include pruning or more advanced methods. 
In general, such approaches are able to provide a large reduction in the size of a trained network, when the model is stored on a disk. 
However, there are some problems when we use such models for inference. 
They are caused by high computation time of sparse computing. 
Another branch of methods uses different matrix-based approaches in neural networks. Thus, there are methods based on the usage of Toeplitz-like structured matrices in  \cite{Lu:LR:2016} or different matrix decomposition techniques: low-rank decomposition
\cite{Lu:LR:2016}, TT-decomposition (Tensor Train decomposition)  \cite{novikov15tensornet,garipov16ttconv}. Also \cite{Bengio:URNN:2016} proposes a new type of RNN, called uRNN (Unitary Evolution Recurrent Neural Networks).

In this paper, we analyze some of the aforementioned approaches.
The material is organized as follows. 
In Section~\ref{sec:languagem}, we give an overview of language modeling methods and then focus on respective neural networks approaches. Next we describe different types of compression. In Section~\ref{sec:prunquant}, we consider the simplest methods for neural networks compression like pruning or quantization. In Section~\ref{sec:lr}, we consider approaches to compression of neural networks based on different matrix factorization methods. Section~\ref{sec:tt} deals with TT-decomposition. Section~\ref{results} describes our results and some implementation details.
Finally, in Section \ref{sec:conclusion}, we summarize the results of our work.


\section{Language modeling with neural networks}\label{sec:languagem}

Consider the language modeling problem. We need to compute the probability of a sentence or
sequence of words $(w_1, \ldots, w_T)$
in a language $L$.
\begin{multline}
    \Prob{\left(w_1, \ldots, w_T\right)} = 
    \Prob{\left(w_1, \ldots, w_{T-1}\right)}
    \Prob{\left(w_T | w_1, \ldots, w_{T-1}\right)} = \\
   = \prod_{t = 1}^{T}\Prob{\left(w_t | w_1, \ldots, w_{t-1}\right)}
\end{multline}

The use of such a model directly would require calculation
$\Prob{\left(w_t | w_1, \ldots, w_{t-1}\right)}$ and in general it is too difficult due to a lot of computation steps.
That is why a common approach features computations with a fixed value of $N$ and approximate (1) with $\Prob{\left(w_t | w_{t-N}, \ldots, w_{t-1}\right)}.$
This leads us to the widely known $N$-gram models~\cite{Jelinek:SMfSR:1997,Kneser:ibfmlm:1995}. It was very popular approach until the middle of the 2000s. A new milestone in language modeling had become the use of recurrent neural networks \cite{Schmid:LSTM:1997}.
A lot of work in this area was done by Thomas Mikolov \cite{Mikolov:2007}.

Consider a recurrent neural network,  RNN, where $N$ is the number of timesteps, $L$ is the number of recurrent layers, $x_{\ell}^{t-1}$ is the input of the layer $\ell$ at the moment $t$.  
Here $t \in \{1,\ldots,N\}$, $\ell \in \{1, \ldots, L\}$, and $x_{0}^{t}$ is the embedding vector. We can describe each layer as follows: 

\begin{align}
    \label{RNN_one_layer} z_{\ell}^t = & W_{\ell}x_{\ell-1}^{t} + V_{\ell}x_{\ell}^{t-1} + b_l   \\
    x_{\ell}^t = & \sigma(z_{\ell}^t),
\end{align}
where $W_{\ell}$ and $V_{\ell}$ are matrices of weights and $\sigma$ is an activation function. 
The output of the network is given by
\begin{equation}
    \label{output_layer}
    y^t = \softmax\left[W_{L+1}x_{L}^t + b_{L+1}\right].
\end{equation}
Then, we define
\begin{gather}
\Prob{\left(w_t | w_{t-N}, \ldots, w_{t-1}\right)} = y^t.
\end{gather}


While $N$-gram models 
even with not very big $N$ require a lot of space due to the combinatorial explosion, neural networks can learn some representations of words and their sequences without memorizing directly all options.

Now the mainly used variations of RNN are designed to solve the problem of decaying gradients
~\cite{Gradient_flow:Hochreiter:2001}. The most popular variation is Long Short-Term Memory (LSTM)
 \cite{Schmid:LSTM:1997} and Gated Recurrent Unit (GRU)  \cite{GRU:Cho:2014}. Let us describe one layer of LSTM: 

\begin{align}
    i_{\ell}^t = & \: \sigma\left[W_l^i x_{l-1}^{t} + V_l^i x_{l}^{t-1} + b_{l}^{i}\right] 
    & \text{input gate}  \\ 
    f_{\ell}^t = & \: \sigma\left[W_l^f x_{l-1}^{t} + V_l^f x_{l}^{t-1} + b_{l}^{f}\right] 
     & \text{forget gate}  \\
    c_{\ell}^t = & \: f_l^t \cdot c_l^{t-1} + i_l^t \tanh\left[W_l^c x_{l-1}^t + U_l^c x_l^{t-1} + b_l^c\right] 
     & \text{cell state} \\
    o_{\ell}^t = & \: \sigma\left[W_l^o x_{\ell-1}^{t} + V_l^o x_{l}^{t-1} + b_{l}^{o}\right] 
     & \text{output gate}  \\ 
    x_{\ell}^t = & \: o_{\ell}^t \cdot \tanh[c_l^t],
\end{align}
where again $t \in \{1,\ldots,N\}$, $\ell \in \{1, \ldots, L\}$, $c_{\ell}^t$ is the memory vector at the layer $\ell$ and time step $t$. The output of the network is given the same formula \ref{output_layer} as above. 

Approaches to the language modeling problem based on neural networks are
efficient and widely adopted, but still require a lot of space.
In each LSTM layer of size $k\times k$ we have 8 matrices of size $k \times k$. Moreover, usually the first (or zero) layer of such a network is an embedding layer that maps word's vocabulary number to some vector. And we need to store this embedding matrix too. Its size is $n_{vocab}\times k$, where $n_{vocab}$ is the vocabulary size.
Also we have an output softmax layer with the same number of parameters as in the embedding, i.e. $k \times n_{vocab}$. 
In our experiments, we try to reduce the embedding size and to decompose softmax layer as well as hidden layers. 

We produce our experiments with compression on standard PTB models. There are three main benchmarks: Small, Medium and Large LSTM models \cite{Zaremba:RNNReg:2014}. But we mostly work with Small and Medium ones. 

\section{Compression methods}
\subsection{Pruning and quantization}\label{sec:prunquant}

In this subsection, we consider maybe not very effective but still useful techniques. 
Some of them were described in application to audio processing~
\cite{Han:DeepCompressing:2016} or image-processing \cite{PrunImage:Molchanov:2016,Image:Svachenko:2017}, but for language modeling this field is not yet well described.

Pruning is a method for reducing the number of parameters of NN. In Fig~1.~(left), we can see that  
usually the majority of weight values are concentrated near zero. It means that such weights do not provide a valuable contribution in the final output. We can set some threshold and then remove all connections with the weights below it  from the network. After that we retrain the network to learn the final weights for the remaining sparse connections.

\begin{figure}[h]
\small{\caption{Weights distribution before and after pruning
 }}
\begin{minipage}[h]{0.47\linewidth}
\center{\includegraphics[width = 60mm]{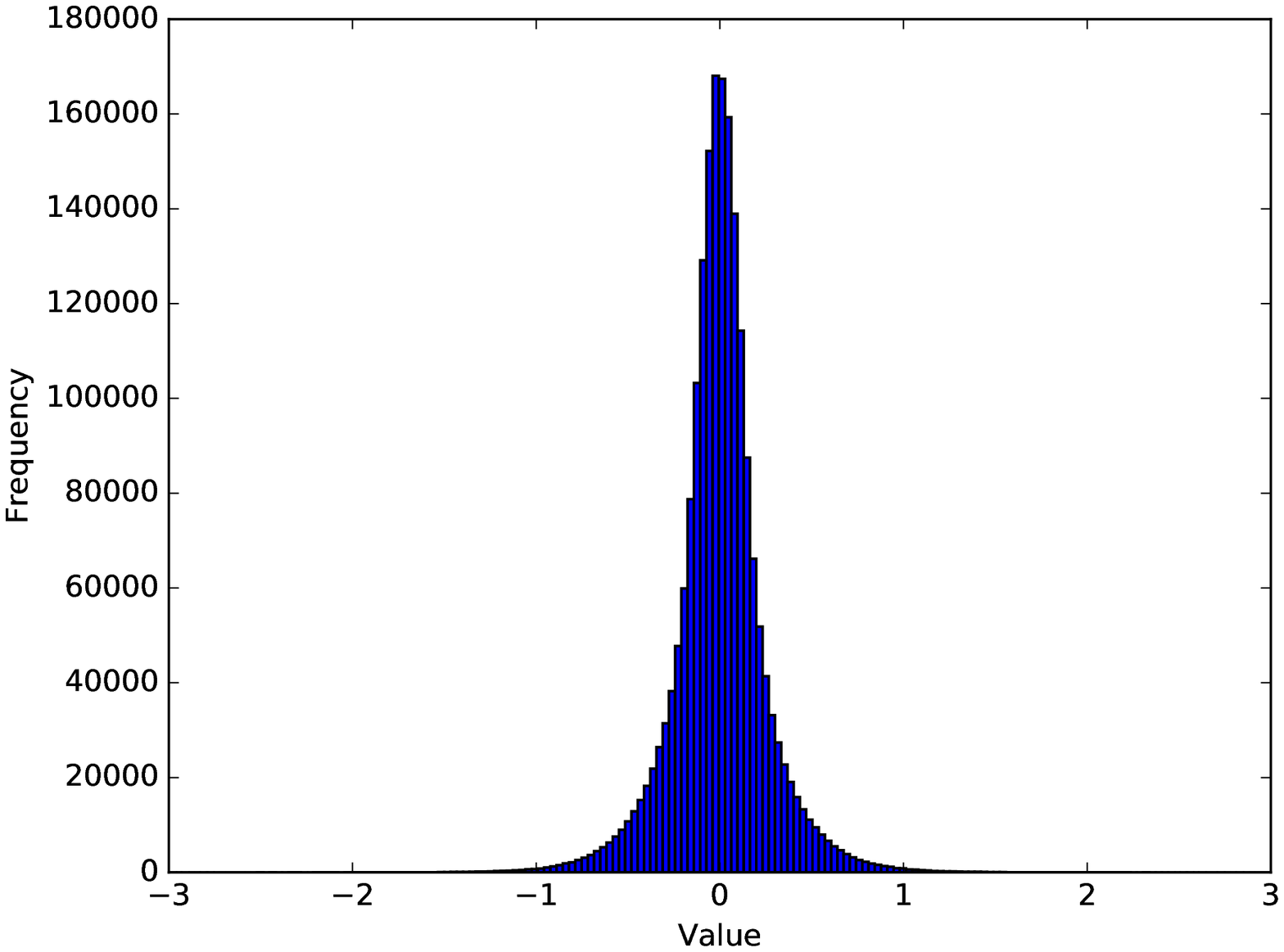}} 
\end{minipage}
\hfill
\begin{minipage}[h]{0.47\linewidth}
\center{\includegraphics[width = 60mm]{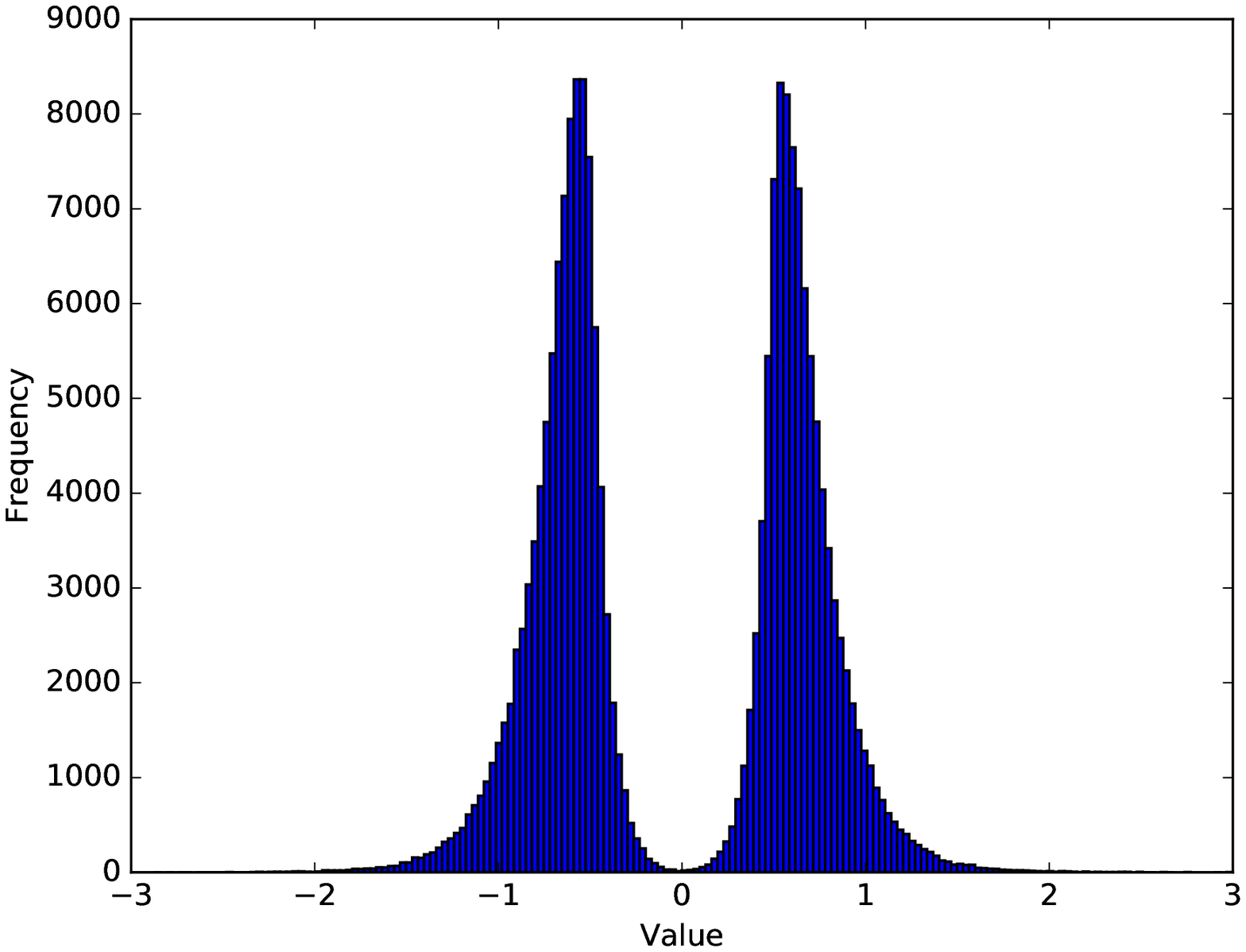}} 
\end{minipage}
\end{figure}

Quantization is a method for reducing the size of a compressed neural network in memory. We are compressing each float value to an eight-bit integer representing the closest real number in one of 256 equally-sized intervals within the range.

Pruning and quantization have common disadvantages since training from scratch is impossible and their usage is
quite laborious.  
In pruning the reason is mostly lies in the inefficiency of sparse computing.
When we do quantization, we store our model in an 8-bit representation, 
but we still need to do 32-bits computations. It means that we have not advantages using RAM. At least until we do not use the tensor processing unit (TPU)
that is adopted for effective 8- and 16-bits computations.

\subsection{Low-rank factorization}\label{sec:lr}

Low-rank factorization represents more powerful methods. 
For example, in \cite{Lu:LR:2016}, the authors applied it to a voice recognition task. 
A simple factorization can be done as follows:

\begin{equation}
    x_l^t = \sigma\left[W_\ell^aW_\ell^bx_{\ell-1}^t + 
    U_l^aU_l^bx_{\ell}^{t-1} + b_l\right]
\end{equation}
Following \cite{Lu:LR:2016} require $W_l^b = U_{\ell-1}^b$. After this we can rewrite our equation for RNN: 
\begin{align}
    x_l^t & =   \sigma\left[W_l^a m_{l-1}^t + U_{l}^a m_{l}^{t-1} + b_l \right] \\
    m_l^t &  =   U_l^b x_l^t
    \\
    y_t & =  \softmax\left[W_{L+1}m_L^t + b_{L+1}\right]
\end{align}

For LSTM it is mostly the same with more complicated formulas. The main advantage we get here from the sizes of matrices $W_l^a$, $U_l^b$, $U_l^a$. They have the sizes $r \times n$ and $n \times r$, respectively, where the original $W_l$ and $V_l$ matrices have size $n \times n$. With small $r$ we have the advantage in size and in multiplication speed. We discuss some implementation details in Section~\ref{results}.

\subsection{The Tensor Train decomposition}\label{sec:tt}

In the light of recent advances of tensor train approach~\cite{novikov15tensornet,garipov16ttconv}, we have also decided to apply this technique to LSTM compression in language modeling. 

The tensor train decomposition was originally proposed as an alternative and more efficient form of tensor's representation~\cite{TT:Oseledets:11}. 
The TT-decomposition (or TT-representation) of a tensor 
$\vec{A} \in \mathbb{R}^{n_1 \times \ldots \times n_d}$ is the set of matrices
$G_k[j_k]  \in  \mathbb{R}^{r_{k-1} \times r_k}$ , where $j_k = 1, \ldots, n_k$, 
$k = 1,\ldots, d$, and 
$r_0 = r_d = 1$ such that each of the
tensor elements can be represented as
$\vec{A}(j_1, j_2, \ldots , j_d) = G_1[j_1]G_2[j_2] \ldots G_d[j_d]. $
In the same paper, the author proposed to consider the input matrix as a multidimensional tensor and apply the same decomposition to it.  
If we have matrix $A$ of size $N \times M$, we can fix $d$ and 
such $n_1, \ldots, n_d$, $m_1, \ldots, m_d$  that the following conditions are fulfilled:
$\prod_{j = 1}^{d}n_j = N$, $\prod_{i = 1}^{d}m_i = M$. Then we 
reshape our matrix $A$ to the tensor $\vec{A}$ with $d$ dimensions and size 
$n_1m_1 \times n_2m_2 \times \ldots \times n_dm_d$. 
Finally, we can perform tensor train decomposition with this tensor. 
This approach was successfully applied to compress fully connected neural networks~\cite{novikov15tensornet} and for developing convolution TT layer~\cite{garipov16ttconv}. 

In its turn, we have applied this approach to LSTM. Similarly, as we describe it above for usual matrix decomposition, here we also describe only RNN layer.
We apply TT-decomposition to each of the matrices $W$ and $V$ in equation~\ref{RNN_one_layer} and get: 

\begin{equation}
   z_{\ell}^t = \TT(W_i)x_{\ell-1}^{t} + \TT(V_l)x_{\ell}^{t-1} + b_{\ell}.
\end{equation}
Here $\TT(W)$ means that we apply TT-decomposition for matrix $W$. It is necessary to note that even with the fixed number of tensors in TT-decomposition and their sizes we still have plenty of variants because we can choose the rank of each tensor. 

\section{Results}\label{results}

For testing pruning and quantization we choose Small PTB Benchmark. 
The results can be found in Table 1. We can see that we have a reduction of the size with a small loss of quality.

For matrix decomposition we perform experiments with Medium and Large PTB benchmarks. When we talk about language modeling, we must say that the embedding and the output layer each occupy one third of the total network size. It follows us to the necessity of reducing their sizes too. We reduce the output layer by applying matrix decomposition. We describe sizes of \textbf{LR LSTM 650-650} since it is the most useful model for the practical application.
We start with basic sizes for $W$ and $V$, $650 \times 650$, and $10000 \times 650$ for embedding. We reduce each 
$W$ and $V$ down to $650 \times 128$ and reduce embedding down to $10000 \times 128$. The value 128 is chosen as the most suitable degree of 2 for efficient device implementation. We have performed several experiments, but this configuration is near the best.  
Our compressed model, \textbf{LR LSTM 650-650}, is even smaller than \textbf{LSTM 200-200} with better perplexity. The results of experiments can be found in Table 2.

\begin{table}[h]
\begin{center}
\caption{\label{results_pq} Pruning and quantization results on PTB dataset}
\begin{tabular}{c|c|c|c}
\hline
\textbf{ Model } & \textbf{ Size  } & \textbf{ No. of params } & \textbf{ Test PP }\\
\hline
\hline
\text{ LSTM 200-200 (Small benchmark) } & 18.6 Mb & 4.64 M &
117.659 \\
\hline
Pruning output layer 90\% & & \\  w/o additional training & 5.5 Mb & 0.5 M &
149.310 \\
\hline
Pruning output layer 90\% & & \\  with additional training & 5.5 Mb & 0.5 M &
121.123 \\
\hline
Quantization (1 byte per number)  &   4.7 Mb  &  4.64 M & 
118.232 \\
\hline
\end{tabular}
\end{center}
\end{table} 

In TT decomposition we have some freedom in way of choosing internal ranks and number of tensors. 
We fix the basic configuration of an LSTM-network with two 600-600 layers and four tensors for each matrix in a layer. And we perform a grid search through different number of dimensions and various ranks. 

We have trained about 100 models with using the Adam optimizer~\cite{Adam:Kingma:2015}. The average training time for each is about 5-6 hours on GeForce GTX TITAN X (Maxwell architecture), but unfortunately none of them has achieved acceptable quality.
The best obtained result (\textbf{TT LSTM 600-600}) is even worse than \textbf{LSTM-200-200} both in terms of size and perplexity.

\begin{table}[h]
\begin{center}
\caption{\label{results_md} Matrix decomposition results
on PTB dataset}
\begin{tabular}{l| c|c | c |c}
\hline
&
\textbf{ Model}
& \textbf{ Size  } & \textbf{ No. of params   } & \textbf{ Test PP   }\\
\hline
\hline
\text{PTB   }  &
LSTM 200-200 & 18.6 Mb & 4.64 M &
117.659  \\
\cline{2-5}
\text{Benchmarks  } & LSTM 650-650 & 79.1 Mb & 19.7 M &
82.07 \\
\cline{2-5}
& LSTM 1500-1500 & \: 264.1 Mb \: & 66.02 M &
78.29 \\
\hline
Ours & LR LSTM 650-650 & 16.8 Mb & 4.2 M&
92.885 \\
\cline{2-5}
& \text{ TT LSTM 600-600 } & 50.4 Mb & 12.6 M&
168.639 \\
\cline{2-5}
& \text{ LR LSTM 1500-1500 } & 94.9 Mb &  23.72 M&
89.462 \\
\hline
\end{tabular}
\end{center}
\end{table}

\section{Conclusion}\label{sec:conclusion}

In this article, we have considered several methods of neural networks compression for the language modeling problem.
The first part is about pruning and quantization. We have shown that for language modeling there is no difference in applying of these two techniques. 
The second part is about matrix decomposition methods. We have shown some advantages when we implement models on devices since usually in such tasks there are tight restrictions on the model size and its structure.
From this point of view, the model \textbf{LR LSTM 650-650} has nice characteristics. It is even smaller than the smallest benchmark on PTB and demonstrates quality comparable with the medium-sized benchmarks on PTB.

\subsubsection*{Acknowledgements.} This study is supported by Russian Federation President grant MD-306.2017.9. A.V. Savchenko is supported by the Laboratory of Algorithms and Technologies for Network Analysis, National Research University Higher School of Economics.

\bibliographystyle{splncs}


\end{document}